%% file: IEEE_IROS.tex
\documentclass[letterpaper, 10 pt, conference]{ieeeconf}  

\IEEEoverridecommandlockouts                              

\usepackage{graphics} 
\usepackage{epsfig} 
\usepackage{float} 
\usepackage{times} 
\usepackage{amsmath} 
\usepackage{amssymb}  
\usepackage{color}
\input{math_commands.tex}

\usepackage{subcaption}
\usepackage{booktabs}
\usepackage{colortbl}
\usepackage{multirow}
\usepackage{makecell}
\usepackage[table]{xcolor}
\definecolor{grey}{rgb}{0.5,0.5,0.5}

\makeatletter
\let\NAT@parse\undefined
\makeatother
\usepackage{hyperref}

\usepackage{babel,blindtext}
\usepackage{placeins}

\usepackage[T1]{fontenc}
\usepackage{aecompl}

\title{\LARGE \bf
A Two-stage Based Social Preference Recognition in Multi-Agent Autonomous Driving System
}

\author{Jintao Xue$^{*}$, Dongkun Zhang$^{*}$, Rong Xiong, Yue Wang, and Eryun Liu
\thanks{$^*$The authors are equally contributed.}
\thanks{Jintao Xue and Eryun Liu are with the Dept of Information Science and Electronic Engineering, Zhejiang University, Hangzhou, China. 
Dongkun Zhang, Rong Xiong and Yue Wang are with the Dept of Control Science and Engineering, Zhejiang University, Hangzhou, China. Eryun Liu is the corresponding author, eryunliu@zju.edu.cn. 
}
}
\begin{document}
\maketitle 
\thispagestyle{empty}
\pagestyle{empty}

\begin{abstract}
Multi-Agent Reinforcement Learning (MARL) has become a promising solution for constructing a multi-agent autonomous driving system (MADS) in complex and dense scenarios. 
But most methods consider agents acting selfishly, which leads to conflict behaviors. 
Some existing works incorporate the concept of social value orientation (SVO) to promote coordination, but they lack the knowledge of other agents' SVOs, resulting in conservative maneuvers. In this paper, we aim to tackle the mentioned problem by enabling the agents to understand other agents' SVOs. To accomplish this, we propose a two-stage system framework. Firstly, we train a policy by allowing the agents to share their ground truth SVOs to establish a coordinated traffic flow. Secondly, we develop a recognition network that estimates agents' SVOs and integrates it with the policy trained in the first stage. Experiments demonstrate that our developed method significantly improves the performance of the driving policy in MADS compared to two state-of-the-art MARL algorithms.
\end{abstract}

\input{sections/1-intro.tex}
\input{sections/2-related-work.tex}

\input{sections/3-preliminaries.tex}

\input{sections/3-method.tex}

\input{sections/4-results.tex}

\input{sections/5-conclusion.tex}

\addtolength{\textheight}{-2cm}   


\input{sections/6-appendix.tex}



\bibliographystyle{IEEEtran}
\bibliography{IEEE_IROS}

\end{document}

%% file: math_commands.tex
\usepackage{amsmath,amsfonts,bm}

\def\eqref#1{equation~\ref{#1}}

\def\1{\bm{1}}

\DeclareMathAlphabet{\mathsfit}{\encodingdefault}{\sfdefault}{m}{sl}
\SetMathAlphabet{\mathsfit}{bold}{\encodingdefault}{\sfdefault}{bx}{n}

\DeclareMathOperator*{\argmax}{arg\,max}

%% file: sections/1-intro.tex
\section{INTRODUCTION}


The self-driving technology is widely regarded as a means to improve the safety and efficiency of transportation, leading to an increasing number of autonomous vehicles undergoing tests in the context of multi-agent autonomous driving systems (MADSs) \cite{wu2021flow}. However, current MADSs encounter difficulties operating in complex environments, including highway merging and bottleneck situations. In these scenarios, road users' actions affect others' behaviors, and a tiny conflict could lead to the decay of traffic efficiency. We aim to address the challenge and design an efficient and safe MADS in these traffic environments.

\begin{figure}[tpb]. 
\centering 
	\includegraphics[width=1.0\linewidth, height=0.76\linewidth]{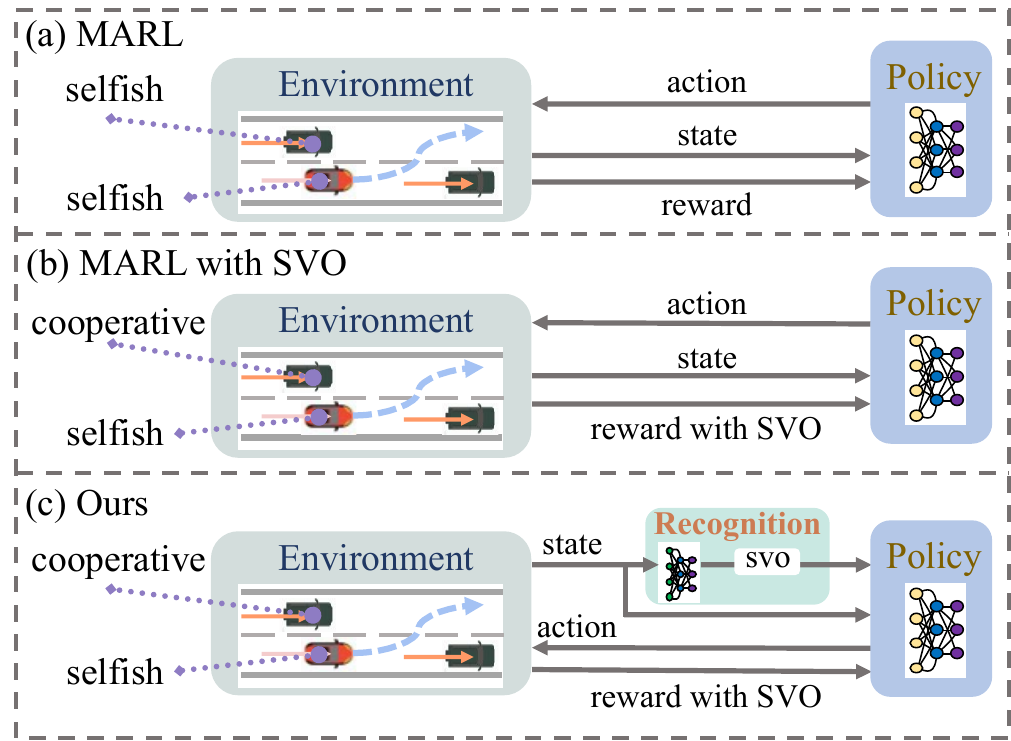}
\caption{Illustration comparing (a) Classical MARL-based approach, (b) MARL with SVO, and (c) our recognition-based policy to have knowledge of other agents' SVOs, thus letting agents make SVO-informed decisions.} \label{fig:illustration}
\vspace{-4mm}
\end{figure}

To this end, we summarize several learning paradigms. The rule-based approaches use manual rules or classical traffic models \cite{treiber2000congested, kesting2007general} but have limitations in complex traffic scenarios. Multi-Agent Reinforcement Learning (MARL) holds great potential and demonstrates positive outcomes \cite{shalev2016safe, wu2021flow, palanisamy2020multi}. As depicted in Fig. \ref{fig:illustration}, most MARL-based MADSs consider agents acting selfishly, producing egoistic behaviors that harm the whole efficiency. To address this issue, several works \cite{ peng2021learning, toghi2021altruistic} introduce the concept of Social Value Orientation (SVO) \cite{schwarting2019social}, which measures the degree of selfishness or altruism of the agent by weighting its rewards with those of others, to promote socially compatible behavior. However, these methods do not know SVOs of other agents and produce conservative behaviors. It is important to note that social preferences, referred to as SVOs, can vary among individuals and significantly affect the interactions between agents \cite{wang2021social}, and neglecting to account for the SVO can negatively impact the safety and efficiency of traffic flow \cite{schwarting2019social}.   

Due to this limitation, in our method, agents are able to understand surrounding agents' SVOs, thereby enabling themselves to make SVO-informed driving, as shown in Fig. \ref{fig:illustration} (c). We start by adopting the MARL approach and incorporating SVO to model MADS. Specifically, we introduce a two-stage training framework. Firstly, we train a policy of actual SVOs, which are the inner parameters of agents to build up a coordinated traffic flow. However, obtaining SVOs from other agents poses a challenge as they are generally considered private information. Additionally, this paper considers the scenario of communication breakdown. 
Hence, in the second stage, we train a recognition policy that can recognize other agents' SVOs and integrate the recognition policy with the policy learned in the first stage. The performance of the two-stage policy is close to that of directly knowing the true SVOs. 
Our experiments explore the effect of varying levels of knowledge among agents on system-level performance metrics, revealing that knowing SVOs leads to more effective and coordinated driving among agents.

To summarize, the main contributions of this paper include the following:
\begin{itemize}
\item A two-stage training procedure by knowing the SVOs of agents in advance to build up a first-stage MADS, then estimating the SVOs to build up the final MADS.  
\item A SVO recognition framework that leverages the power of self-attention models to tackle complex driving environments. This framework seamlessly integrates multiple sources of information to obtain accurate estimations. 
\item Our proposed approach for MADS is validated on a simulation environment and compared against two state-of-the-art MARL-based methods. The evaluation results demonstrate that our method outperforms others in terms of performance.
\end{itemize}

%% file: sections/2-related-work.tex
\section{Related works}
In this section, we review relevant MARL approaches in the context of MADS. We also discuss recent techniques used to generate heterogeneous driving behavior and online parameter estimation approaches for navigation.

\subsection{Deep Reinforcement Learning in MADS}
The application of Multi-Agent Reinforcement Learning (MARL) produces promising outcomes in MADS.
Authors of \cite{shalev2016safe} apply the DRL method to the problem of forming long-term driving strategies while ensuring functional safety, and a hierarchical temporal abstraction is introduced to reduce the variance of the gradient estimation. Palanisamy \textit{et al.} \cite{palanisamy2020multi} provide a simulation platform and a taxonomy of multi-agent learning environments to help further research. Wang \textit{et al.} \cite{wang2020multi} utilize graph attention networks in the navigation setting of MARL for mixed-autonomy cooperation. Liu \textit{et al.} \cite{liu2020platoon} propose a distributed training framework for deep Q-networks to deal with the multi-vehicle platooning problem. \cite{toghi2021cooperative,toghi2022social,toghi2021altruistic} incorporate SVO to acquire socially compliant behavior, and the third paper obtains better results using a multi-agent actor-critic algorithm. Dai \textit{et al.} \cite{daisocially} dynamic change SVOs for interacting agents in each episode. Peng \textit{et al.} \cite{peng2021learning} use the coordination factor to facilitate the coordination of agents at both local and global levels in fully autonomous traffic flow. We follow this approach by incorporating SVO. However, we design a recognition framework to know other agents' SVOs.

\subsection{Heterogeneity of Vehicle Agents}
Considerable methods design heterogeneous agents having diverse driving styles to reflect real-world driving scenes. The intelligent driver model (IDM) \cite{treiber2000congested} and the MOBIL lane-changing model are often combined as a model of human drivers \cite{kesting2007general}. Furthermore, some approaches adjust the IDM-MOBIL model's parameters (e.g., politeness factor) to get different levels of aggressiveness. Saxena \textit{et al.} \cite{saxena2020driving} modify IDM to include a stop-and-go behavior and diverse cooperativeness. Mavrogiannis \textit{et al.} \cite{mavrogiannis2022b} present an algorithm to conduct behavior-rich simulation consisting of egoistic and conservative agents. \cite{wang2021socially, huang2021driving, schwarting2019social, sun2018courteous} follow a weighted cost function, and varied weight metrics characterizes the difference between individuals, and the weights can be manually tuned or apply Inverse Reinforcement Learning (IRL) to learn from real-human data. Schwarting \textit{et al.} \cite{schwarting2019social} also preferably employ SVO in autonomous driving, determining the degree of competitive and prosocial. And many researchers integrate the concept of SVO into their works \cite{peng2021learning, toghi2021altruistic, wang2021socially, crosato2022interaction, zhao2021yield, li2022efficient, buckman2021semi}.
\subsection{Online Parameter Estimation}
Several works estimate social preferences in driving. \cite{hoermann2017probabilistic,bhattacharyya2020online} use online filtering techniques to estimate parameters in IDM. Authors of \cite{le2021lucidgames} use an unscented Kalman filter to iteratively update a Bayesian estimate of other agents' cost function parameters. 
Li \textit{et al.} \cite{li2022efficient} identify other drivers' driving preferences by estimating the SVOs. 
\cite{schwarting2019social, zhao2021yield} estimate the SVO of the agent to improve predictions and prove essential assets for interactive driving. 
Wang \textit{et al.} \cite{wang2021socially} allow agents to infer other road users' characteristics include egoism, courtesy, and confidence. However, these model-based methods assume that the agent fully knows the state transition function or other agents' objective functions.
But what if the function is a black box, or more precisely, the internals of the environment will be unknown to an agent? 
For example, give random discrete estimation values of one objective, and each value needs to go through the black box and get the outcome, which means running a significant number of parallel processes to get the running results, which is costly to call. However, our policy can handle the black-box environment with dense agents as we do not need the objective or transition function.

%% file: sections/3-preliminaries.tex

%% file: sections/3-method.tex
\section{Method}
Based on the discussion above, we propose a two-stage strategy to solve MADS problem using MARL: (i) train a policy to coordinate traffic flow by knowing true SVOs, (ii) adopt a policy to estimate agents' SVOs. For ease of reference, we denote these two policies as the ``decision policy'' and the ``recognition policy''. First, we provide problem definitions of the two policies. We then describe the reward function and state representation used in the environment, then followed by an introduction to the overall network architecture.

\subsection{Decision Policy Training} \label{section:decision-policy}

\subsubsection{Partially Observable Stochastic Game (POSG)} 
We formulate the decision-making processes in MADS using a stochastic game \cite{oliehoek2016concise} by the tuple $G = \left<\mathcal{I}, \mathcal{S}, \mathcal{A}, \mathcal{O}, P, \mathcal{R}, n, \rho_0, \gamma, T\right>$.
$\mathcal{I}$ represents a finite set of $n$ agents, $\mathcal{S}$ represents the state-space of all agents, while $\mathcal{A} = \mathcal{A}_{1} \times \mathcal{A}_{2} \dots \times \mathcal{A}_{n}$, and $\mathcal{O} = \mathcal{O}_{1} \times \mathcal{O}_{2} \dots \times \mathcal{O}_{n}$ denote the joint action, and observation spaces, respectively. 
At a time agent $i \in \mathcal{I}$ receive the observation $o_i : \mathcal{S} \to \mathcal{O}_i$ and take action based on the shared policy $\pi: \mathcal{O}_i \times {\mathcal{A}_i} \to \left[0,1\right]$.
Consequently, the full state changes from s to s’ after all agents
take their actions w.r.t the state transition function $P : \mathcal{S} \times \mathcal{A} \times \mathcal{S} \to \left[0,1\right]$. The rewards denoted by an agent-specific reward functions $\mathcal{R}_i \in \mathcal{R}$ where $\mathcal{R} = \{ R_0, R_1, \dots, R_{n-1} \} $. $\rho_0 $ is the initial state distribution, $\gamma \in (0, 1]$ is the discount factor, and $T$ is the time horizon. 

 \begin{figure*}[tpb] 
 \vspace{2mm}
\centering	
	\includegraphics[width=0.94\linewidth, height=0.45\linewidth]{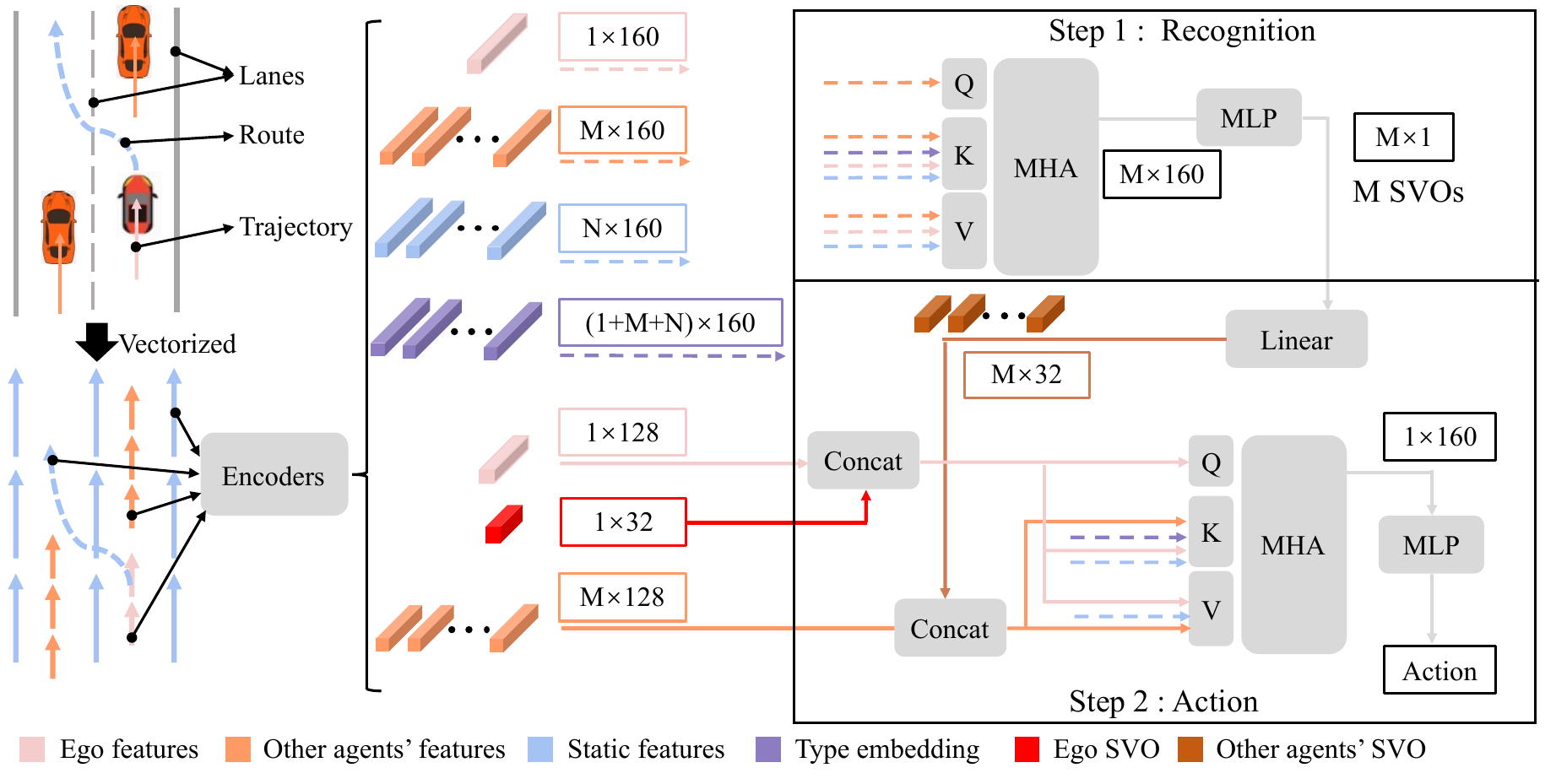}
		\label{network}
\caption{The architecture of our policy consists of encoders that embed the input vectorized information into different features. These features then go through the first multi-head attention (MHA) layer on the upper right, which outputs information about M surrounding agents' SVOs. The resulting features then pass through the second MHA to produce the final output.\label{fig: architecture}} 
\end{figure*}

\subsubsection{Decentralized reward function} 
Our approach leverages the SVO concept from prior research \cite{schwarting2019social, buckman2019sharing, peng2021learning} and integrates it into our framework. By doing so, each agent is able to exhibit varying behaviors, such as aggressive or cooperative, depending on the extent to which they consider other agents' rewards:
\begin{equation} \label{equation:reward-svo}
\begin{gathered}
    R_i = \cos(\frac{\pi}{2} \cdot \phi_i) r_i+ \sin(\frac{\pi}{2} \cdot \phi_i) r^s_i ,\\
    r_{i}^{s} = \frac{\sum_{j\in \mathcal{D}} r^j}{|\mathcal{D}|}, \, \mathcal{D} = \{ j: || \text{Pos}(i) - \text{Pos}(j) || \leq d \} ,   
\end{gathered}
\end{equation}
the reward $r_i$ represents the individual driving performance, which contains metrics such as average speed and a negative reward in case of collision. $r_i^s$ is the average of surrounding agents' utilities, and the $d$ means each agent only perceives information from other agents within a certain Euclidean distance. $\phi_i \in [0, 1]$ is the SVO of agent $i$ and remains constant throughout each episode.

\subsubsection{Objective Function} We adopt the decentralized learning method to solve the optimization objective of POSG:
\begin{align}
    \pi^* = \argmax_{\pi} \mathbb{E}_{\pi} [\sum_{t=0}^{T}\gamma^{t} R_i(s_t, a_t)],  \quad i \in \mathcal{I} ,   \label{equation:object-posg}
\end{align}
 where $\pi: \mathcal{O}_i \times {\mathcal{A}_i} \to \left[0,1\right]$, which uses experiences of all agents to learn a strategy with sharing of policy network parameters. Due to unique observations and indices by agents, diverse behaviors emerge, aligning with concepts presented in \cite{terry2020revisiting}. 


\subsection{Recognition Policy Training} \label{section:recognition policy}
According to \ref{section:decision-policy}, we can train a decision policy under the assumption of knowing other SVOs. However, it is more reasonable to consider SVO as an inner parameter that can not be directly observed. This section presents a policy to recognize surrounding agents' SVOs, coupled with the trained decision policy. 
Our training approach involves directly fitting the actual and the predicted value of SVOs. 
Given a set of observations, including static and vehicular features, our task is to predict the values of SVOs for surrounding agents. To generate training data, we run the POSG with the trained decision policy and store the resulting observations. We denote SVO as $\phi$ and define the mean square error (MSE) loss function as the average of squared differences between the actual and the predicted values of SVOs:
\begin{align}
    \mathcal{L}_{reg} = \frac{1}{N\sum_{i=0}^N {M_i}}\sum_{i=0}^N\sum_{j=0}^{M_i} {(\phi_{ij} - \hat{\phi_{ij}})^2}.
  \label{equation:loss-regression}
\end{align}
Given N agent's observations, each agent needs to estimate the values of $\phi$ for its neighboring agents. $M_i$ denote the number of surrounding agents for the $i$-th agent, and $\hat{\phi_{ij}}$ represent the estimated value of the $j$-th agent's $\phi$ by the $i$-th agent.

\subsection{State Space Representation} \label{section:state-space}
\subsubsection{State Space for decision policy} \label{section:state-space-action}
To improve computational and memory efficiency, we implement a vectorized representation strategy known as VectorNet \cite{gao2020vectornet}. The state space contains static and vehicular set $\chi=\left\{\chi^s, \chi^v\right\}$, and elements in $\chi^s$ and $\chi^v$ are sets of points containing corresponding features. For the static set containing road centerlines, sidelines, and routes, $\chi^s = \left\{centerline, sideline, route\right\} = \left\{e^s_0, e^s_1, \dots, e^s_i, \dots\right\}$, where $e^s_i = \left\{ \xi_0, \xi_1, \dots, \xi_j, \dots \right\}, i \in \chi^s$. $\xi_j = \left[ p_j, \phi_i, i, j \right]$, in which $p_j = (x, y, heading)$ is the pose of point $j$ in element $i$ and $\phi_i$ is the lane width of element $i$. 
For the vehicular set cover poses and velocities of n agents, $\chi^v = \left\{e^v_0, e^v_1, \dots, e^v_{n-1}\right\}$, in which $e^v_i = \left\{ \xi_0, \xi_1, \dots, \xi_{horizon} \right\}$, $i \in \chi^v $, and $\xi_j = \left[p_j, \phi_i, i, j \right]$, $j \in e_i^v$, where $p_j = (x, y, heading, speed)$ and $\phi_i$ denotes the SVO of agent $i$. In practice, setting the agent's trajectory horizon to 10 achieves good driving performance without exceeding computational resources.
\subsubsection{State space for recognition policy}
Almost the same as mentioned in \ref{section:state-space-action}, the state space for recognition comprises static and vehicular elements. However, the recognition policy does not require knowledge of the true SVOs, hence $\xi_j = [p_j,i,j]$, where $j \in e^v_i$.

\subsection{Reward Function Design}
The driving task involves multiple attributes when defining the reward function. These attributes include factors such as comfort and compliance with traffic regulations. To address this, we design a reward function that provides continuous incentives for driving fast while imposing penalties for catastrophic failures. These failures encompass collisions with other agents, deviations from the designated driving zone, and excessive deviation from the global path.

\subsection{Policy Architecture} \label{section:method-policy-architecture}
The entire network architecture is depicted in Figure \ref{fig: architecture}, including encoders, recognition, and decision components. The process starts with encoding observations of the ego agent into high-level features and embedding diverse features into a uniform dimensional space as described in \ref{section:architecture-deepset}. The recognition policy then focuses on M surrounding agents and receives their driving behavior information, which is embedded into M SVOs as outlined in \ref{section:architecture-recognition}. Finally, the decision policy receives SVOs, including the ground truth of the ego agent, and outputs the final action detailed in \ref{section:architecture-decision}.

\subsubsection{DeepSet Encoder} \label{section:architecture-deepset}
Based on theorem 2 in \cite{zaheer2017deep}, the representation element $e\subset \chi$, where $\chi = \left\{ \chi^s, \chi^v \right\}$, requires a function that preserves the adjacency between elements and is permutation-invariant to the order of objects in the element. Hence, the propagation function $f$ is defined as follows:
\begin{equation}
   f(e) = \rho \left( \sum_{\xi \in {e}} \varphi(\xi) \right).
   \label{equation:architecture-deepset}
\end{equation}
We obtain features at the element level by transforming the nodes $\xi \in e$ into a representation $\varphi(\xi)$. The sum of representations is processed using the $\rho$ network defined by Multi-Layer Perception (MLP) network. The DeepSet architecture enables us to extract features at the polyline level while keeping the number of parameters relatively small.

\subsubsection{Recognition Policy Architecture} \label{section:architecture-recognition}
Through DeepSet, the input features are embedded into a 160-dimensional space and categorized into three groups: ego, other agents, and static elements. The multi-head attention (MHA) layer is then applied, with features of $M$ surrounding agents serving as queries $Q_r = [q_r^1, q_r^2, \dots, q_r^M] \in \mathbb{R}^{M\times {d_k}}$, where $d_k$ is the dimension of the key vectors, set to 160. All features are keys $K_r$ or values $V_r$. Following the work in \cite{carion2020end,scheel2022urban}, we incorporate an additional type embedding into the keys to allow the model to attend to values based on object types. Depicting the complete computation process as follows:
\begin{eqnarray}
    \Phi_r = \operatorname{tanh}(\operatorname{Decoder}(\operatorname{MultiHead}(Q_r, K_r, V_r))),
    \label{equation:architecture-attn1}
\end{eqnarray}
where MultiHead composes several Attention operations, which calculate the weighted sum of the values using the dot-product of queries and keys. The outputs of $\operatorname{Attention}$ are concatenated and then transformed using a linear layer to obtain the final representation. $\operatorname{Attention}$ is defined as:

\begin{eqnarray}
    \operatorname{Attention}(Q_r, K_r, V_r)=\operatorname{softmax}\left(\frac{Q_r K_r^T}{\sqrt{d_k}}\right) V_r.
    \label{equation:architecture-attn}
\end{eqnarray}
 For simplicity, we use an MLP as the decoder function. Via the decoder and $\operatorname{tanh}$, the output of the MHA layer is projected to a single-dimensional space and gets the final recognition result, $\Phi_r = [\phi_r^1, \phi_r^2, \dots, \phi_r^M] \in \mathbb{R}^{M\times {1}}$.

\subsubsection{Decision Policy Architecture}\label{section:architecture-decision}
As mentioned in \ref{section:state-space}, the decision policy's definition of vehicular elements $\xi^v$ differs from that of the recognition policy. Hence, another DeepSet-based encoder embeds the vehicular elements into a 128-dimensional feature space. Next, we project the self-true SVO and the estimated SVOs of M surrounding agents into a 32-dimensional space using a linear layer, and the resulting features are concatenated with vehicular element features and passed through another MHA network, $\operatorname{MultiHead}(Q_a, K_a, V_a)$. Unlike the recognition policy, we only use a single query $Q_a = [q_a] \in \mathbb{R}^{1\times {160}}$ by features of ego agent. Finally, the output of MHA is decoded into the action $a \in \mathcal{A}$.

%% file: sections/4-results.tex
\section{Experiments}
In this section, we pursue to answer several questions. (1) Can our recognition-based method achieve superior system-level performance? (2) Can our recognition framework successfully estimate agents' SVOs? Additionally, we investigate the factors that affect the accuracy of recognition. 

\subsection{Experimental Setup}
We utilize the Universe simulator \cite{zhang_Universe_2023} to simulate bottleneck and merging scenarios. To model the motion of the vehicles in the simulator, we employ the Kinematic Bicycle Model and utilize a closed-loop proportional-integral-derivative (PID) controller to translate the actions into low-level steering and acceleration control signals. To allow for a continuous representation of action, we use the $a = [speed, heading] \in \mathbb{R}^2$ notation, with values bounded by the range of $[-1, 1]$, then mapped to the speed range of $[0, 6m/s]$ and the steering angle range of $[-\pi/4, \pi/4]$, respectively.
During the training phase, in each episode, the agents are randomly spawned within a range of 8 to 20, and we randomly initialize their spawn points, global paths, and SVOs. We assumed vehicles have optimal conditions for map information, perception, localization, and control to focus on planning during the simulation. In the testing phase, we fix the number of agents at 20. All experiments are performed on a computer with an Intel i9-12900KF CPU and NVIDIA GeForce RTX 3090. 

\subsection{Training}
We use Independent Policy Learning (IPL) \cite{tan1993multi} for training the decision policy. To train the IPL within single-agent reinforcement learning, we utilize Soft-Actor-Critic (SAC) \cite{haarnoja2018soft}.
We train our recognition policy using the supervised learning approach. In particular, we execute the decision policy trained from the first stage in the environment by acquiring the true value of SVOs and subsequently use generated offline data to train the recognition policy. We use the Adam optimizer \cite{kingma2015adam} to optimize both policies.

\subsection{Metrics} \label{section:metric}
Our experiments are evaluated based on measures of both efficiency and safety. We evaluate safety by calculating the percentage of an episode resulting in accidents, including the frequency of departures from the designated driving zone, collisions into the wrong lane, and driving too far from the global path. 
For brevity, we denote the above three types of accidents as ``Crash''. To measure the recognition accuracy at the system level, we consider the mean deviation error between the multi-agent recognition values and the corresponding true values.

\begin{table}[ht] \label{appendix-table:traffic-simulation}

\caption{The table presents the percentage of various metrics (defined in section \ref{section:metric}) for the bottleneck and merge scenarios, along with the performance of our proposed method indicated by a ``$^\dagger$".}

\begin{subtable}[h]{\linewidth}
\centering
\begin{tabular}{b{2.5cm}<{\centering} c c c c}
    \toprule 
    \multirow{2}{*}[-0.5em]{\makecell[c]{\vspace{0.1cm}Methods \\}} & \multicolumn{3}{c}{\texttt{Bottleneck}} \\
    \cline{2-4} \\ 
    & Success ($\uparrow$) & Crash ($\downarrow$) & Speed ($\uparrow$) \\
    \cline{1-4}  \vspace{-10mm}\\ 
    {MACAD \cite{palanisamy2020multi}} & 76.1 $\pm$ 0.3 & 24.1 $\pm$ 0.6  & 75.1 $\pm$ 0.2  \\
    {CoPO \cite{peng2021learning}}   & 80.3 $\pm$ 0.6  & 20.7 $\pm$ 1.1 & 74.5 $\pm$ 0.3   \\
    {TrueSVO}  & 83.1 $\pm$ 0.4   & {16.9 $\pm$ 1.0}  & 76.3 $\pm$ 0.2    \\
     {Recog}$^\dagger$    & {82.3 $\pm$ 0.3}   & {17.3 $\pm$ 1.0}  & 76.0 $\pm$ 0.1   \\
    \bottomrule 
\end{tabular} 
\end{subtable}
\begin{subtable}[h]{\linewidth}
\centering
\begin{tabular}{b{2.5cm}<{\centering} c c c c}
    \vspace{0.8mm}
    \multirow{2}{*}[-0.5em]{\makecell[c]{\vspace{0.1cm}Methods \\}} & \multicolumn{3}{c}{\texttt{Merge}} \\
    \cline{2-4} \\ 
    & Success ($\uparrow$) & Crash ($\downarrow$) & Speed ($\uparrow$) \\
    \cline{1-4}  \vspace{-10mm}\\ 
    {MACAD {\cite{palanisamy2020multi}}}      & 66.1 $\pm$ 0.6  & 34.0 $\pm$ 0.9  & 55.0 $\pm$ 0.2   \\
    {CoPO \cite{peng2021learning}}  & 69.3 $\pm$ 0.5  & 30.7 $\pm$ 0.9 & 54.9 $\pm$ 0.2    \\
    {TrueSVO} & 
    82.9 $\pm$ 0.4  & 17.2 $\pm$ 0.6 &  60.0 $\pm$ 0.1  \\
    {Recog}$^\dagger$ & 81.8 $\pm$ 0.3  & 18.4 $\pm$ 0.7 & 59.6 $\pm$ 0.1  \\
    \bottomrule 
\end{tabular} 
\end{subtable}
\vspace{-2mm} \label{table:performance}
\end{table}

\subsection{Performance of Multi-agent Driving System}
\begin{figure}
\vspace{2mm}
\centering

   \includegraphics[width=0.94\linewidth]{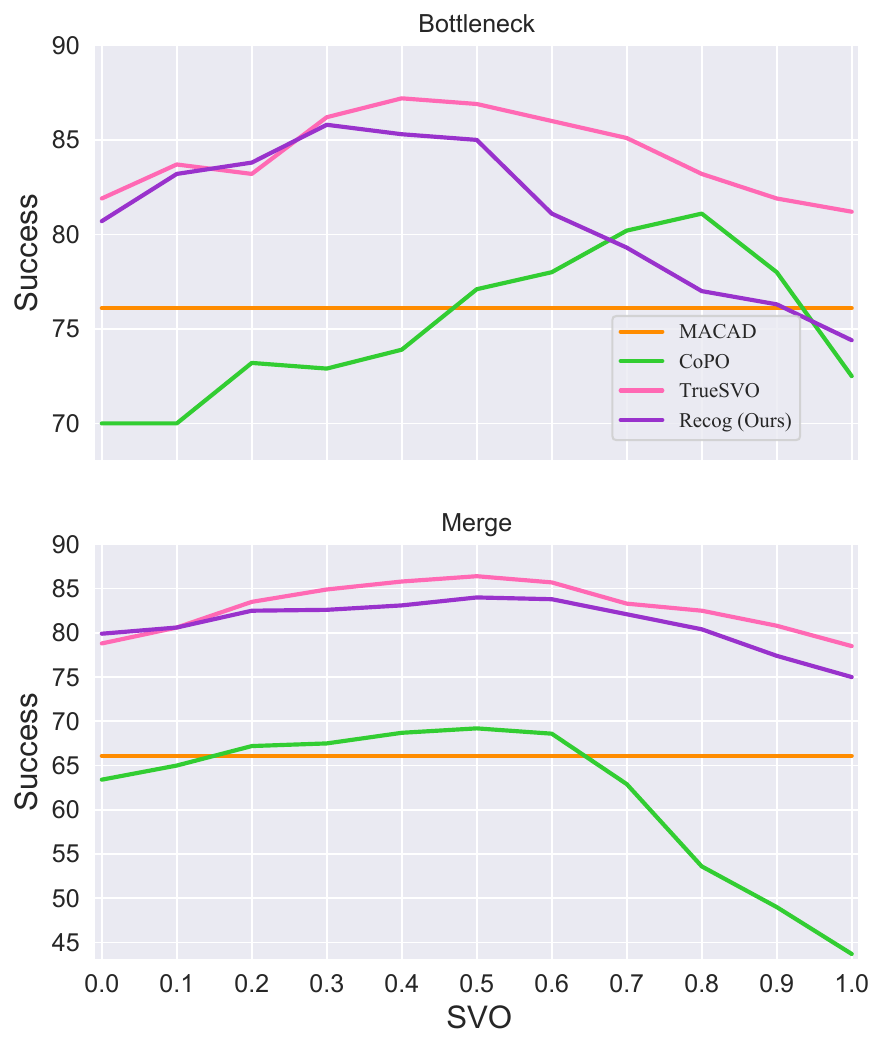}
   \vspace{-1mm}
   \caption{Success rates. The figure shows the percentage of success rates in the bottleneck and merge. We assign a fixed SVO from 0 to 1 at regular intervals. All agents in traffic flows are given the same SVO. Each is evaluated for 200 episodes. MACAD actually does not have the concept of SVO, it is used here as a reference line.}\label{fig:fix_svo}
\vspace{-3mm}
\end{figure}

We compare our proposed approach with two MARL-based baselines, MACAD \cite{palanisamy2020multi} and CoPO \cite{peng2021learning}. MACAD is an approach that considers each agent aiming to maximize its reward. While CoPO incorporates SVO to promote coordination among agents at both local and global levels but does not know other agents' SVO. Our approach, where the recognition policy estimates the SVOs and passes them to the decision policy referred to as Recog. We also take our decision policy as a comparison method denoting TrueSVO, where the TrueSVO receives true value of SVOs directly. Evaluation results shown in Table. \ref{table:performance} and Fig. \ref{fig:fix_svo}, The MACAD approach produces individual egoistic behaviors, results in lower performance. While CoPO achieves better coordination and higher success rates in MADS with random SVOs but has lower average speeds across two scenarios. 
The Recog approach offers insights into other agents' driving styles and performs better than MACAD and CoPO, but due to estimation errors, its performance is lower compared to TrueSVO.
TrueSVO outperforms other approaches in all metrics, indicating that sharing driving attitude information leads to more effective and coordinated MADS.

\vspace{-1mm}

\subsection{Recognition Accuracy}

\begin{figure*}[!htb]
\centering
\vspace{2mm}
  \includegraphics[width=0.98\linewidth, height=0.28\linewidth]{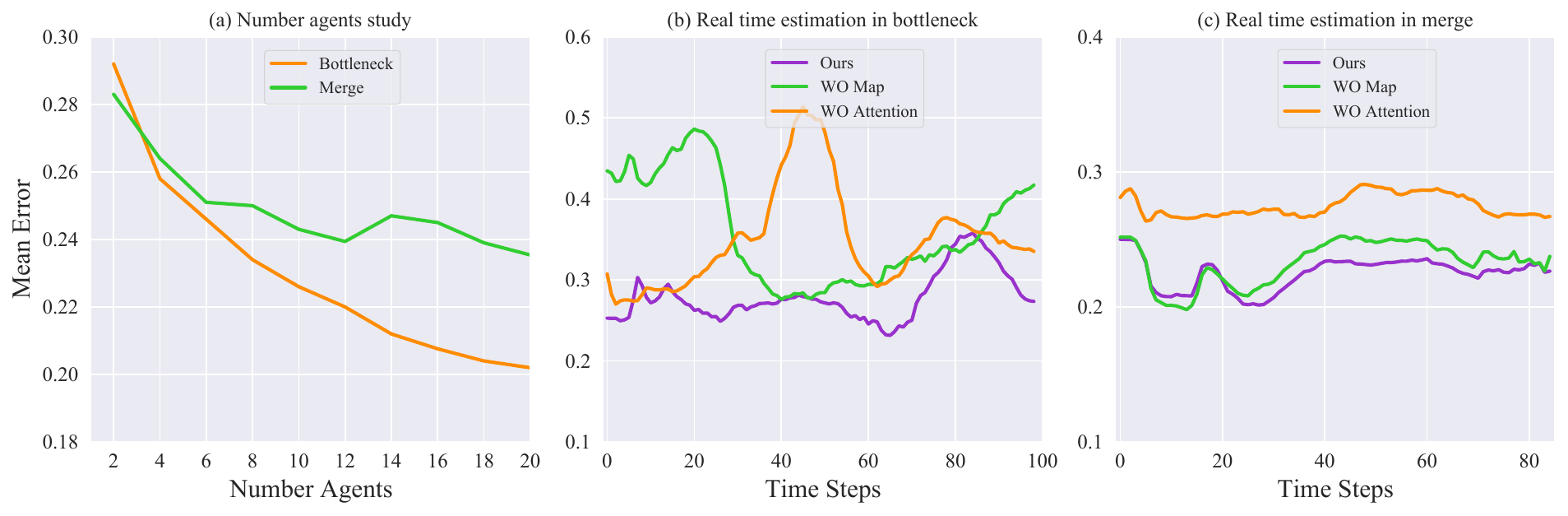}
 \caption{Influences of accuracy. In (a), we change the number of agents in the bottleneck and merge. In (b) and (c), we change the architecture of our network, including removing the road information (WO Map) or attention model (WO Attention). We use the mean deviation error mentioned in \ref{section:metric} as metric. Each is evaluated for 200 episodes.} \label{fig:ablation}
 \vspace{-2mm}
\end{figure*} 

\begin{figure}
\centering
\vspace{1mm}
\begin{subfigure}[b]{0.5\textwidth}
   \includegraphics[width=0.97\linewidth]{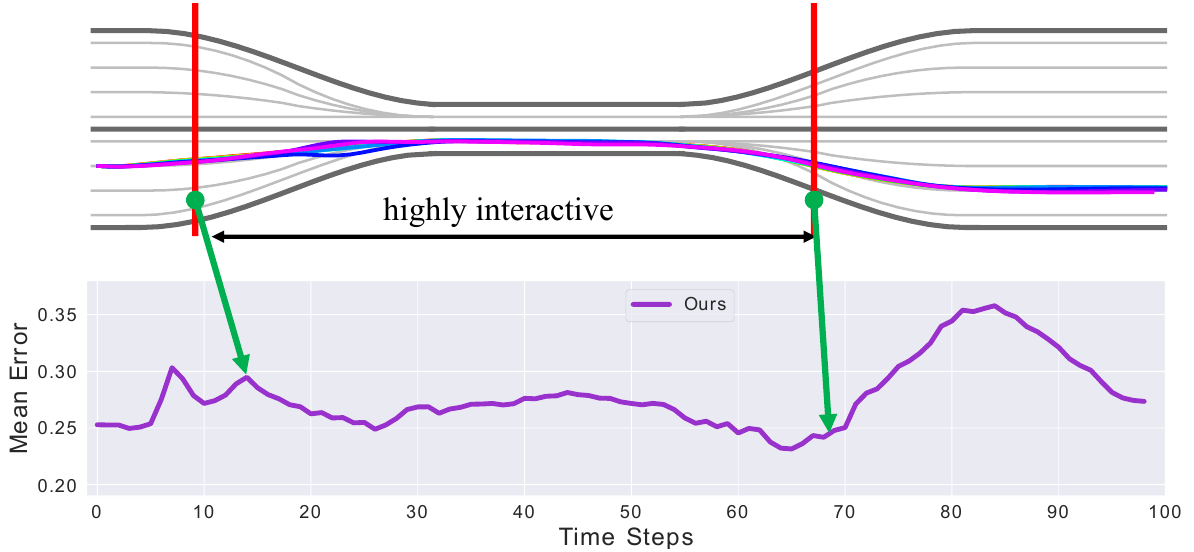}
   \vspace{-1mm}
   \caption{Bottleneck.}\label{fig:road_bottleneck}
\end{subfigure}
\vspace{2.5mm}
\begin{subfigure}[b]{0.5\textwidth}
   \includegraphics[width=0.97\linewidth]{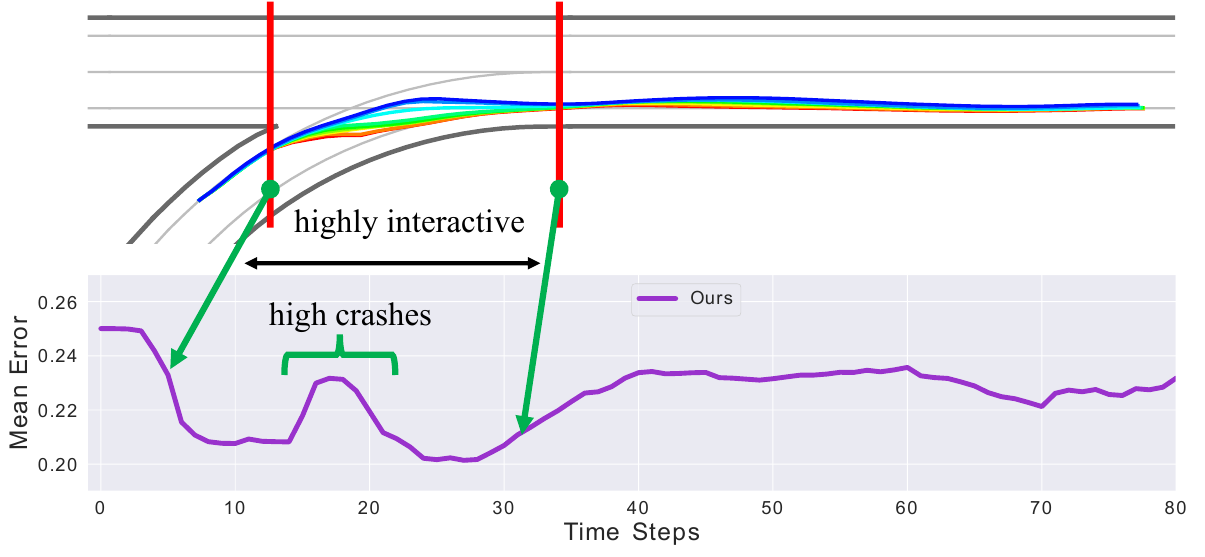}
   \vspace{-1mm}
   \caption{Merge.}\label{fig:road_merge}
\end{subfigure}
\vspace{-5mm}
\caption{Visualization of scenarios and one vehicle's trajectories. }\vspace{-5.5mm}\label{fig:road}
\end{figure}
\subsubsection{Impact of Agents Number}
As depicted in Figure \ref{fig:ablation}, it can be observed that our policy demonstrates better precision
convergence as the number of agents increases.
When the number of agents decreases, the influence of individual driving behaviors on other agents reduces, and the demand for coordinated behaviors among agents is lower, making it more difficult for our strategy to identify the characters of the agents from the interaction. While the number increase, 
more interactions exist, thus the policy shows a rising performance of estimation of SVOs.

\subsubsection{Highly Interactive Period}
Fig. \ref{fig:road} presents a visualization of the scenes and the roll-out trajectories of one agent, with time period markings of high-interaction environments. The result shows our policy achieves better accuracy in the highly interactive period. In the bottleneck, the recognition error rises at the beginning, as the agents exhibit little driving information. Furthermore, during times 15 and 70, the agents navigate through a highly interactive environment as each agent slows down into a narrow lane and interacts with others, exhibiting diverse behaviors, allowing the policy to estimate more accurately. In the merge scenario, the error declines from steps 0 to 30 because agents going straight need to avoid merging agents, providing more information for the recognition policy. However, the time steps around 18 are with high incidences of crashes, which causes increasing estimation error. In the second half of the time, the agents experience a less competitive road structure and exhibit more homogeneous behaviors, leading to an increase in recognition error. 

\begin{table}[ht]
\caption{Label Interpretation.\label{table:labels}}
\centering
\begin{tabular}{c|c|c|c} 
\hline
         & \begin{tabular}[c]{@{}l@{}}trajectories\end{tabular} & \begin{tabular}[c]{@{}l@{}} \quad road\\structures\end{tabular} & \begin{tabular}[c]{@{}l@{}}attention\\network\end{tabular} \\
\hline
Without attention & \checkmark &  &  \\
Without map (with attention)& \checkmark &  & \checkmark \\
With attention and map (Ours) & \checkmark & \checkmark & \checkmark\\
\hline
\end{tabular} 
\end{table}

\subsubsection{Importance of Attention-based Model}
Fig. \ref{fig:ablation} (b) and (c) show how the map information, such as lane and boundary data, impacts the recognition accuracy of SVOs. Table \ref{table:labels} explains the labels used in these figures. 
Though performing well in the merge, the recognition policy without map information performs badly in the bottleneck, due to the fact that the bottleneck has a more complex road structure that lasts for a longer time period, and the lack of map information makes it difficult for the policy to estimate the SVOs of the surrounding agents accurately. As for the policy without attention, it only knows the agents' trajectories and performs worst in both scenarios. 
Our method utilizes the attention model that combines map information with agents' trajectories to help to know the surrounding environments better and get the best performance. Although our method's accuracy decreases in the bottleneck during the last thirty time steps, this can be attributed to the fact that agents are in a less interactive environment, making it more challenging to estimate their characteristics accurately. These findings suggest that combining multiple sources of information can lead to more accurate SVOs recognition and enhance the effectiveness of autonomous driving systems.

%% file: sections/5-conclusion.tex
\section{Conclusions}

This paper focuses on the challenge of designing a safe and efficient MADS and introduces a novel social preference recognition framework to handle complex driving environments. The framework can integrate multiple sources of information to achieve more accurate social preference recognition. We propose a two-stage method for MADS, which comprises a recognition policy and a decision policy that are seamlessly integrated. We evaluate our method on two complex scenarios, namely bottleneck and merge, and compare its performance with other MARL-based methods. 
The results demonstrate that sharing SVOs can lead to better performance of MADS, highlighting the effectiveness of our approach.
